\documentclass[10pt,twocolumn,letterpaper]{article}

\usepackage{wacv}
\usepackage{times}
\usepackage{epsfig}
\usepackage{graphicx}
\usepackage{amsmath}
\usepackage{amssymb}
\usepackage{booktabs}
\usepackage[ruled,vlined, linesnumbered]{algorithm2e}
\def\wacvPaperID{1577} % Enter the WACV Paper ID here

\wacvalgorithmstrack   % Uncomment this line if you are submitting to the Algorithms Track.
\wacvfinalcopy % *** Uncomment this line for the final submission

\ifwacvfinal
\usepackage[breaklinks=true,bookmarks=false]{hyperref}
\else
\usepackage[pagebackref=true,breaklinks=true,colorlinks,bookmarks=false]{hyperref}
\fi

\begin{document}

%%%%%%%%% TITLE
\title{Learning Less Generalizable Patterns with an Asymmetrically Trained Double Classifier for Better Test-Time Adaptation}

\author{Thomas Duboudin$^1$\\
% Univ Lyon, Ecole Centrale de Lyon, CNRS, \\ INSA Lyon, Univ Claude Bernard Lyon 1, \\ Univ Louis Lumière Lyon 2, LIRIS, UMR5205\\
% 69134 Ecully, France\\
% {\tt\small firstname.name@ec-lyon.fr}
% For a paper whose authors are all at the same institution,
% omit the following lines up until the closing ``}''.
% Additional authors and addresses can be added with ``\and'',
% just like the second author.
% To save space, use either the email address or home page, not both
\and
\hspace{-0.5cm}
Emmanuel Dellandréa$^1$\\
\and 
\hspace{-0.5cm}
Corentin Abgrall$^2$\\
\and 
\hspace{-0.5cm}
Gilles Hénaff$^2$\\
\and
\hspace{-0.5cm}
Liming Chen$^1$\\
\and 
Univ Lyon, Ecole Centrale de Lyon, CNRS, \\ 
INSA Lyon, Univ Claude Bernard Lyon 1, \\ 
Univ Louis Lumière Lyon 2, LIRIS, UMR5205,\\
69134 Ecully, France\\
{\tt\small firstname.name@ec-lyon.fr}
\and Thales LAS France, \\
78990 Élancourt, France\\
{\tt\small firstname.name@fr.thalesgroup.com}
}

\maketitle
\thispagestyle{empty}

%%%%%%%%% ABSTRACT
\begin{abstract}
Deep neural networks often fail to generalize outside of their training distribution, in particular when only a single data domain is available during training. While test-time adaptation has yielded encouraging results in this setting, we argue that, to reach further improvements, these approaches should be combined with training procedure modifications aiming to learn a more diverse set of patterns. Indeed, test-time adaptation methods usually have to rely on a limited representation because of the shortcut learning phenomenon: only a subset of the available predictive patterns is learned with standard training. In this paper, we first show that the combined use of existing training-time strategies, and test-time batch normalization, a simple adaptation method, does not always improve upon the test-time adaptation alone on the PACS benchmark. Furthermore, experiments on Office-Home show that very few training-time methods improve upon standard training, with or without test-time batch normalization. We therefore propose a novel approach using a pair of classifiers and a shortcut patterns avoidance loss that mitigates the shortcut learning behavior by reducing the generalization ability of the secondary classifier, using the additional shortcut patterns avoidance loss that encourages the learning of samples specific patterns. The primary classifier is trained normally, resulting in the learning of both the natural and the more complex, less generalizable, features. Our experiments show that our method improves upon the state-of-the-art results on both benchmarks and benefits the most to test-time batch normalization.
\end{abstract}
\newpage

%%%%%%%%% BODY TEXT
\section{Introduction}

Deep neural networks performance falls sharply when they are confronted, at test-time, with data coming from a different distribution, or domain, than the training one. A change in lighting, sensor, weather conditions, or geographical location can result in a dramatic performance drop \cite{hoffman2018cycada, beery2018recognition, degrave2021ai}. Such environmental changes are commonly encountered when an embedded network is deployed in the wild, and exist in such diversity that it is impossible to gather enough data to cover all possible domain shifts. This lack of cross-domain robustness prevents the widespread deployment of deep networks in safety critical applications.\\ 

Domain generalization algorithms have been investigated to mitigate the test-time performance drop by modifying the training procedure. Contrary to the domain adaptation research field, in which unlabeled samples of the target distribution are available at training time \cite{ganin2015unsupervised, tzeng2017adversarial}, no information about the target domain is assumed to be known in domain generalization. Most of them assume to have access to data coming from several identified different domains, and try to create a domain invariant representation by finding common predictive patterns \cite{li2018domain, moyer2018invariant, carlucci2019domain, li2018learning, krueger2020out, huangRSC2020}. However, such an assumption is quite generous and in many real-life applications one does not have access to several data domains but only a single one. As a result, a number of methods study single-source domain generalization \cite{wang2021learning, shi2020informative, zhao2020maximum, zhang2022exact, nam2021reducing}. A majority of methods were however found to perform only marginally better than the standard training procedure when the evaluation is done rigorously on several benchmarks \cite{gulrajani2021in, zhang2022nico++}. Another recent paradigm, called test-time adaptation, proposes to use a normally trained network, and adapt it with a quick procedure at test-time, using only a batch of unlabeled target samples. This paradigm yielded promising results in the domain generalization setting \cite{you2021test, yang2022test} because they alleviate the main challenges of domain generalization: the lack of information about the target domain, and the requirement to be simultaneously robust in advance to every possible shift.\\

Test-time adaptation methods however suffer from a drawback that limits their adaptation capability and that can only be corrected at training-time. Indeed, using a standard training procedure, only a subset of predictive patterns is learned, corresponding to the most obvious and efficient ones, while the less predictive patterns are disregarded entirely \cite{singla2021understanding, hermann2020shapes, pezeshki2020gradient, hermann2020origins, shah2020pitfalls, beery2018recognition, geirhos2020shortcut}. This apparent flaw, named shortcut learning, originates from the gradient descent optimization \cite{pezeshki2020gradient} and prevents a test-time method to use all the available patterns. The combination of a training-time patterns' diversity seeking approach with a test-time adaptation method may thus lead to improved results. In this paper, we show that the combined use of test-time batch normalization, a simple test-time adaptation method, with the state-of-the-art single-source domain generalization methods (that are often designed to discover normally unused patterns) does not systematically yield increased results on the PACS benchmark \cite{pacs} in the single-source setting. Similar experiments on Office-Home \cite{officehome} yield a similar result, with only few methods performing better than the standard training procedure. We thus propose a new method, namely L2GP, which encourages a network to learn new predictive patterns rather than exploiting and refining already learned ones and demonstrate its effectiveness on both the PACS and the Office-Home benchmarks. To find such patterns, we propose to look for predictive patterns that are less generalizable than the naturally learned ones through a secondary classifier endowed with a shortcut avoidance loss, thereby leading to learning semantically different patterns. These less generalizable patterns match the ones normally ignored because of the simplicity bias of deep networks that promotes the learning of a representation with a high generalization capability \cite{huh2021low, galanti2022sgd}. Our method requires two classifiers added to a features extractor and trains them asymmetrically, using a data-dependant regularization, \textit{e.g.}, shortcut avoidance loss, that slightly encourages memorization rather than generalization by learning batch specific patterns, \textit{i.e.} patterns that lower the loss on the running batch but with a limited effect on the other batches of data.\\

To summarize, our contribution is threefold: 
\begin{itemize}

    \item To the best of our knowledge, we are the first to investigate the effect of training-time single-source methods on a test-time adaptation strategy. We show that it usually does not increase performance and can even have an adverse effect.
    \item We apply, for the first time, several state-of-the-art single-source domain generalization algorithms on the more challenging and rarely used Office-Home benchmark and show that very few yield a robust cross-domain representation.
    \item We propose an original algorithm to learn a larger than usual subset of predictive features and show that it yields results over the existing state-of-the-art with the combination of test-time batch normalization.     
\end{itemize}

\begin{figure}
\includegraphics[scale=1.12]{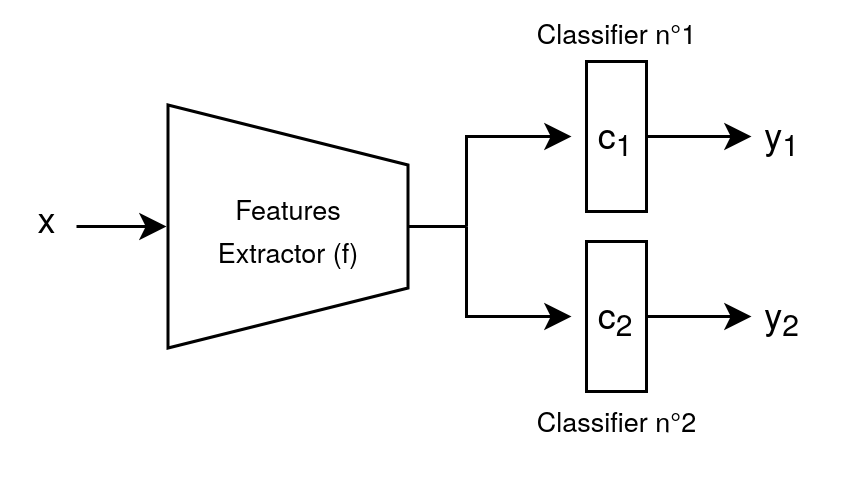}
\caption{Schema of our bi-headed architecture. The naming convention is the same as the one used in algorithm \ref{algo}.}
\label{fig:archi}
\end{figure}

\section{Related Works}

\subsection{Single-Source Domain Generalization}

Most domain generalization algorithms require several identified domains to enforce some level of distributional invariance. Because this is an unrealistic hypothesis in some situations (such as in healthcare or defense related tasks), methods were developed to deal with a domain shift issue with only one single domain available during training. Some of them rely on a domain shift invariance hypothesis. A commonly used invariance hypothesis is the texture shift hypothesis: a lot of domain shifts are primarily textures shifts, and using style transfer based data augmentation will improve the generalization, whether it is explicitly by training a model on stylized images \cite{wang2021learning, jackson2019style} or implicitly in the internal representation of the network \cite{zhang2022exact, nam2021reducing}. Such methods are limited to situations where it is indeed a shift of the hypothesized nature that is encountered. Others wish to learn a larger set of predictive patterns to make the network more robust should one or several training-time predictive patterns be missing at test-time. Volpi \etal \cite{ada} and Zhang \etal \cite{zhao2020maximum} propose to incrementally add adversarial images crafted to maximize the classification error of the network to the training dataset. These images no longer contain the original obvious predictive patterns which forces the learning of new patterns. These strategies are inspired by adversarial training methods \cite{huang2015learning, kurakin2016adversarial} that were originally designed to improve adversarial robustness in deep networks. Wang \etal \cite{wang2021learning} used a similar approach in an online fashion, without the impractical ever-growing training dataset, and combined it with a style augmentation approach. Huang \etal \cite{huangRSC2020} and Shi \etal \cite{shi2020informative} used a dropout \cite{dropout} based strategy to prevent the network from relying only on the most predictive patterns by muting the most useful channels or mitigating the texture bias. These methods were evaluated in the single-source setting on several benchmarks, among which the very common PACS dataset.

\subsection{Test-Time Adaptation}

Test-time adaption has emerged as a promising paradigm to deal with domain shifts. By waiting to gather information about the target domain, in the shape of an unlabeled batch of samples (or even a single sample), it alleviates the main drawbacks of training-time domain generalization methods: the lack of information about the target domain, and the necessity to simultaneously adapt to all possible shifts. The simplest test-time adaptation strategy consists of replacing the training-time statistics in the batch normalization layers with the running test batch statistics. This is now a mandatory algorithm block for almost all methods \cite{nado2020evaluating, benz2021revisiting, you2021test, hu2021mixnorm, schneider2020improving}. This strategy was originally designed to deal with test-time images corruptions but proved to be efficient in a more general domain shift setting \cite{you2021test, yang2022test}. In a situation where samples of a test batch cannot be assumed to come from the same distribution, workarounds requiring a single sample were developed by mixing test-time and training-time statistics \cite{you2021test, yang2022test, hu2021mixnorm, schneider2020improving}, or by using data augmentation \cite{hu2021mixnorm}. Some solutions, such as the work of Yang \etal \cite{yang2022test} or Wang \etal \cite{wang2021tent} further rely on test-time entropy minimization to remove inconsistent features from the prediction. Finally, Zhang \etal \cite{zhang2021memo} quickly adapt a network to make consistent predictions between different augmentations of the same test sample. All these strategies rely on a model trained with the standard training procedure. 

\section{Method}

\begin{algorithm*}
% \small
\SetAlgoLined
\textbf{Method specific hyperparameters:} \\
- weight for the shortcut avoidance loss $\alpha$ \\
- step size used for the gradient perturbation $lr_{+}$ \\
\textbf{Networks:} \\
- features extractor $f$, and its weights $W$ (ResNet18 without its last linear layer) \\
- first classifier $c_1$ (single linear layer)\\
- second classifier $c_2$ (single linear layer)\\
 \While{training is not over}{
  sample 2 batches of data $\{(x_i,y_i), i=0...N-1\}, \{(\tilde{x}_i,\tilde{y}_i), i=0...N-1\}$\\
  calculate the cross-entropy loss $\mathcal{L}$ on the first batch for both branches on the original weights $W$:\\
  $\mathcal{L}(f, c_1) = \frac{1}{N} \sum_i\mathcal{L}[c_1(f(W, x_i)), y_i]$\\
  $\mathcal{L}(f, c_2) = \frac{1}{N} \sum_i\mathcal{L}[c_2(f(W, x_i)), y_i]$\\
  calculate the gradient of the cross-entropy loss $\mathcal{L}$ w.r.t $W$ on the first batch:\\ 
  $\nabla_W \mathcal{L} = \nabla_W \frac{1}{N} \sum_i \mathcal{L}[c_2(f(W, x_i)), y_i]$\\
  add the perturbation to the running weight $W$, and track this addition in the computational graph:\\
  $W_{+} = W + lr_{+} \nabla_W \mathcal{L}$ \\ % {||\nabla_W \mathcal{L}||_2^2}$\\
  calculate the shortcut avoidance loss on the second batch: \\
  $\mathcal{L}_{sa}(f, c_2) = \frac{1}{N} \sum_i ||c_2(f(W, \tilde{x}_i))-c_2(f(W_{+}, \tilde{x}_i))||_1$\\
  update all networks to minimize $\mathcal{L}_{total}(f, c_1, c_2) =  \frac{1}{2} (\mathcal{L}(f, c_1) + \mathcal{L}(f, c_2)) + \alpha \mathcal{L}_{sa}(f, c_2)$
  }
  \textbf{At test-time:} use $c_1 \circ f$ (discard $c_2$) combined with test-time batch normalization
\caption{Learning Less Generalizable Patterns (L2GP)}
\label{algo}
\end{algorithm*}

Because of the simplicity bias of deep networks, the "hidden" patterns we are searching for are patterns that are predictive, but with a lower generalization capacity than the naturally learned ones. We wish to learn both the "hidden" and the naturally learned patterns, as they are not necessarily spurious, and therefore should not be systematically ignored. If we are able to reach a low task loss value on a certain batch of data, without witnessing a similar decrease on another batch of the same distribution, it means that the patterns learned are both predictive and generalize poorly, which is precisely the patterns we are looking for. We therefore wish to move in the parameters space in a direction that lowers the loss on the running batch and keep the loss at its current value on a different batch. This would require a direct optimization of the update direction, which is memory and time consuming. Instead, to implement this idea practically, we rely on a relaxation of the problem. The complete procedure is available in algorithm \ref{algo}. We first add a lightweight modification to the standard network architecture, illustrated in figure \ref{fig:archi}, that is compatible with a wide range of networks and tasks: a secondary prediction layer is added after the features extractor and next to the original one, \eg for a classification task, using a ResNet \cite{he2016deep}, we simply add a single fully connected layer that takes as input the features map after the average pooling operation (as with the original classification layer). We then train both branches of the architecture with the same classification cross-entropy loss on the running batch (algo. \ref{algo}, lines 10-12). The secondary branch has an additional shortcut avoidance loss: we compare the second prediction on a new batch with and without applying a single cross-entropy gradient ascent step to the features extractor's weights, where the gradient is computed using the original running batch (algo. \ref{algo}, lines 13-18). For a network trained using the standard training procedure, these two predictions differ greatly as the features learned in the gradient step generalize from a batch to another and thus result in an important loss increase. By training the features extractor (alongside with the secondary classifier) to reduce the gap between both predictions, and with the classification loss, we are driving it to learn weights that are predictive for the running classification batch, but that have a low effect on the predictions of another batch. The first branch is needed to learn the original patterns. The first classifier will use every available features at disposal, including those learned by the secondary branch, while the secondary will only favor less simple ones. Only the first classifier is therefore used at test-time, the second one can be discarded.\\

During the evaluation, we use test-time batch-normalization (TTBN). This method has been chosen because of its simplicity and its wide range of applicability. Instead of using the exponential average training mean and standard deviation in the batch-normalization layers, we first calculate the statistics on the running test batch, use them to update an exponential average of the target statistics, as in \cite{nado2020evaluating, benz2021revisiting, you2021test, hu2021mixnorm, schneider2020improving}, before using this estimate to normalize the features. A correct target statistics approximation can be reached only if all samples encountered at test-time come from the same data distribution. This is a realistic scenario for applications like autonomous driving, in which the data distribution is not expected to change over the course of a few consecutive images, but less so for networks made available online as an API where samples from a completely different distribution can be sent at the same time. Several methods \cite{you2021test, hu2021mixnorm} provide ways to circumvent this issue if needed.

\section{Experiments and discussion}

\subsection{Baselines for comparison}

We compare our approach with the standard training procedure (expected risk minimization, ERM), with several methods designed for single-source domain generalization \cite{wang2021learning, zhang2022exact, nam2021reducing, ada, zhao2020maximum, shi2020informative}, with a method designed to reduce the shortcut learning phenomenon in deep networks \cite{pezeshki2020gradient} and a multi-source domain generalization algorithm that does not explicitly require several training domains \cite{huangRSC2020}. These baselines were selected because they yield state-of-the-art results, are representative of the main ideas in the single-source domain generalization research community, and because they have a publicly available implementation. This was a necessity as the original works' results were given without any test-time adaptation and trained models were not provided. Our experiments are conducted on the PACS (7 classes, 4 domains, around 10k images in total), and the Office-Home (65 classes, 4 domains, around 15k images in total) benchmarks. PACS has been often used in the single-source setting, but not Office-Home. For the PACS datasets, on which most of the baselines were tested, we use the original works' hyper-parameters. For the Office-Home datasets, we used the hyper-parameters of the multi-source setting if available. If the methods did not have quantitative hyper-parameters, such as EFDM \cite{zhang2022exact} with the choice of mixing-layers depths, we used the ones proposed for the PACS experiments. Finally, for the remaining works, we conducted a simple hyper-parameters search using a single training-test domains pair, and transferred them as is to the other pairs of the same training domain.

\subsection{Experimental setup}

For all the methods and benchmarks, we use the data augmentation described in \cite{huangRSC2020} (random resized crops, color jitter, random horizontal flips, random grayscale). The models selected for the test are those with the best validation accuracy, to avoid a target domain information leak. For a particular domain used for training, 90\% of the dataset is used for training and the remaining 10\% for validation. The test set is obtained using another domain dataset entirely. Experiments were conducted with a ResNet18 \cite{he2016deep} trained for 100 epochs, with the stochastic gradient descent, a learning rate of $1e-3$, a batch size of $64$, a weight decay of $1e-5$, and a Nesterov momentum of $0.9$. After 80 epochs, the learning rate is divided by 10. We chose to use the same common hyper-parameters for all baselines to precisely measure the effect of the training procedure modifications rather than the influence of a perhaps better than usual hyper-parameter. This change of hyper-parameters may yield some small inconsistencies between the results reported in the original works and ours. The gradient ascent learning rate is set to $1.0$ and the $\alpha$ weight for the shortcut avoidance loss to $1.0$ as well, for all the experiments, that is for all the training-test pairs on both the PACS and the Office-Home datasets. The exponential average momentum used in the batch normalization layers at test-time is set to $0.1$.

\subsection{Results and analysis}

\begin{table*}
\begin{center}
% \footnotesize
\begin{tabular}{|c|c|c|c|c|}
\hline
  & \multicolumn{2}{|c|}{\textit{without TTBN}} & \multicolumn{2}{|c|}{\textit{with TTBN}}\\
\hline
Method & Avg. Val. Acc. & Avg. Test Acc. & Avg. Val. Acc. & Avg. Test Acc.\\
\hline
\multicolumn{5}{|c|}{\textbf{PACS dataset}}\\
\hline
ERM & $ 96.8 \pm  0.4$ & $ 52.0 \pm  1.9$  & $ 97.4 \pm  0.3$ & $ 66.1 \pm  1.1$ \\
\hline
RSC \cite{huangRSC2020} & $ 97.7 \pm  0.4$ & $ 54.3 \pm  1.8$ & $97.2 \pm  0.2$ & $ 58.7 \pm  1.6$ \\ 
\hline 
InfoDrop \cite{shi2020informative} & $ 96.6 \pm  0.3$ & $ 53.4 \pm  2.0$ & $ 95.9 \pm  0.3$ & $ 65.5 \pm  1.0$ \\
\hline
ADA \cite{ada} & $ 96.9 \pm  0.8$ & $ 55.9 \pm  2.9$ & $ 96.6 \pm  1.1$ & $ 66.5 \pm  1.2$ \\
\hline
ME-ADA \cite{zhao2020maximum} & $ 96.7 \pm  1.3$ & $ 54.7 \pm  3.1$ & $ 96.5 \pm  0.9$ & $ 66.7 \pm  2.0$ \\
\hline
EFDM \cite{zhang2022exact}& $ 96.9 \pm  0.5$ & $59.6 \pm  2.3$ & $ 97.5 \pm  0.5$ & $ \mathbf{71.3 \pm  1.0}$ \\
\hline
SagNet \cite{nam2021reducing} & $ 97.2 \pm  0.7$ & $ 57.9 \pm  2.9$ & $ \mathbf{97.8 \pm  0.7}$ & $ 62.4 \pm  1.8$ \\
\hline 
L.t.D \cite{wang2021learning} & $ 97.9 \pm  1.0$ & $\mathbf{ 59.9 \pm  2.7}$ & $ 97.6 \pm  0.7$ & $ 66.3 \pm  1.5$ \\
\hline
Spectral Decoupling \cite{pezeshki2020gradient} & $ 95.9 \pm  0.4$ & $ 52.9 \pm  2.6$  & $ 96.2 \pm  0.7$ & $ 66.7 \pm  1.1$ \\
\hline
\textbf{L2GP (ours)} & $ \mathbf{98.6 \pm  0.2}$ & $ 56.1 \pm  2.7$ & $ 96.4 \pm  0.3$ & $ \mathbf{71.3 \pm  0.6}$ \\
\hline
\multicolumn{5}{|c|}{\textbf{Office-Home dataset}}\\
\hline
ERM & $ 82.0 \pm  0.8$ & $ 52.0 \pm  0.8$ & $ 81.6 \pm  1.1$ & $ 52.6 \pm  0.6$ \\
\hline
RSC \cite{huangRSC2020} & $ 80.9 \pm  0.4$ & $ 49.2 \pm  0.7$ & $ 80.2 \pm  0.5$ & $ 48.9 \pm  0.7$ \\
\hline 
InfoDrop \cite{shi2020informative} & $ 76.4 \pm  0.8$ & $ 45.9 \pm  0.5$ & $ 77.1 \pm  0.7$ & $ 46.4 \pm  0.6$ \\
\hline
ADA \cite{ada} & $ 81.2 \pm  2.6$ & $ 50.4 \pm  0.9$ & $ 80.3 \pm  2.0$ & $ 50.0 \pm  0.7$ \\
\hline
ME-ADA \cite{zhao2020maximum}  & $ 78.9 \pm  1.4$ & $ 49.8 \pm  0.6$  & $ 81.4 \pm  1.2$ & $ 50.0 \pm  0.7$ \\
\hline
EFDM \cite{zhang2022exact} & $ 82.9 \pm  0.5$ & $ 52.8 \pm  0.6$ & $ 83.3 \pm  1.0$ & $ 53.3 \pm  0.5$ \\
\hline
SagNet \cite{nam2021reducing} & $ 81.5 \pm  1.5$ & $ 51.9 \pm  0.7$ & $ 81.1 \pm  1.1$ & $ 51.8 \pm  0.9$ \\
\hline 
L.t.D \cite{wang2021learning}& $ 81.0 \pm  1.2$ & $ 50.9 \pm  0.7$ & $ 81.7 \pm  2.7$ & $ 51.2 \pm  0.8$ \\
\hline
Spectral Decoupling \cite{pezeshki2020gradient} & $ 83.8 \pm  0.7$ & $ 52.5 \pm  0.5$ & $ 82.5 \pm  0.6$ & $ 53.2 \pm  0.3$ \\
\hline
\textbf{L2GP (ours)} & $ \mathbf{84.0 \pm  0.6}$ & $ \mathbf{53.4 \pm  0.6}$ & $ \mathbf{83.8 \pm  0.5}$ & $\mathbf{ 54.5 \pm  0.3}$\\
\hline
\end{tabular}
\end{center}
\caption{Performances of our approach and comparison with the state-of-the-art.}
\label{table:main_results}
\end{table*}

The main results of our paper are available in Table \ref{table:main_results}. They were obtained as follows: first, for all the 12 distinct pairs of training and test domains, we calculate the average and the standard deviation of the validation and test accuracies over 3 runs (because the effect of the network's initialization on the test accuracy is greater than usual in a test-time domain shift situation). The reported numbers are the average over all distinct pairs of the pairwise average accuracies $\pm$ the average over all distinct pairs of the pairwise standard deviation (as we are interested in the randomness of the initialization rather than the variation of accuracies between training-test pairs). Used alongside test-time batch normalization, our method reaches a performance similar to that of EFDM \cite{zhang2022exact} on the PACS datasets, but exceeds it on the Office-Home datasets. When test-time batch normalization is not used, our method remains state-of-the-art on the Office-Home dataset but falls behind the style transfer based methods on the PACS dataset from a noticeable margin. Besides, our approach also benefits the accuracy on the validation sets.\\

We observe a completely different behavior between experiments on PACS and Office-Home. While all the existing methods improve upon the standard training procedure (ERM) on PACS, only EFDM, spectral decoupling \cite{pezeshki2020gradient}, and our method yield better results on Office-Home. Likewise, while always positive, the effect of the test-time batch normalization is much more noticeable on PACS than on Office-Home. Furthermore, it is interesting to notice that the performance gain due to the test-time batch normalization is highly dependant on the training-time method used, with the gain being the highest when our approach or ERM is used and only reaching a result closely similar to ERM or below in most of the other cases. We hypothesize that the domain shifts of the PACS datasets are mostly textures shifts, while they are not for the Office-Home datasets. This would explain why test-time batch normalization yield an large improvement on the PACS benchmark, as the simple use of test-time statistics, that encode textures \cite{benz2021revisiting}, is enough to significantly bridge the domain gap and why the methods reaching the highest results \cite{zhang2022exact, nam2021reducing, wang2021learning} in the usual setting (without test-time batch normalization) are all style transfer based methods. As our approach is not related to style transfer in any way, we are able to reach a higher accuracy on Office-Home than other existing works. Regarding the effect of different training-time methods, we hypothesize that the gain is related to whether or not the method is not learning a more diverse set of patterns but rather weighting differently patterns that would be naturally learned (style transfer based methods, for instance, essentially grant a higher importance to shape-based patterns rather than texture-based patterns, but not necessarily learn new patterns) which would explain why several methods that improve upon ERM without test-time batch normalization only perform precisely as well once it is used.\\

\begin{figure*}
\includegraphics[width=\textwidth]{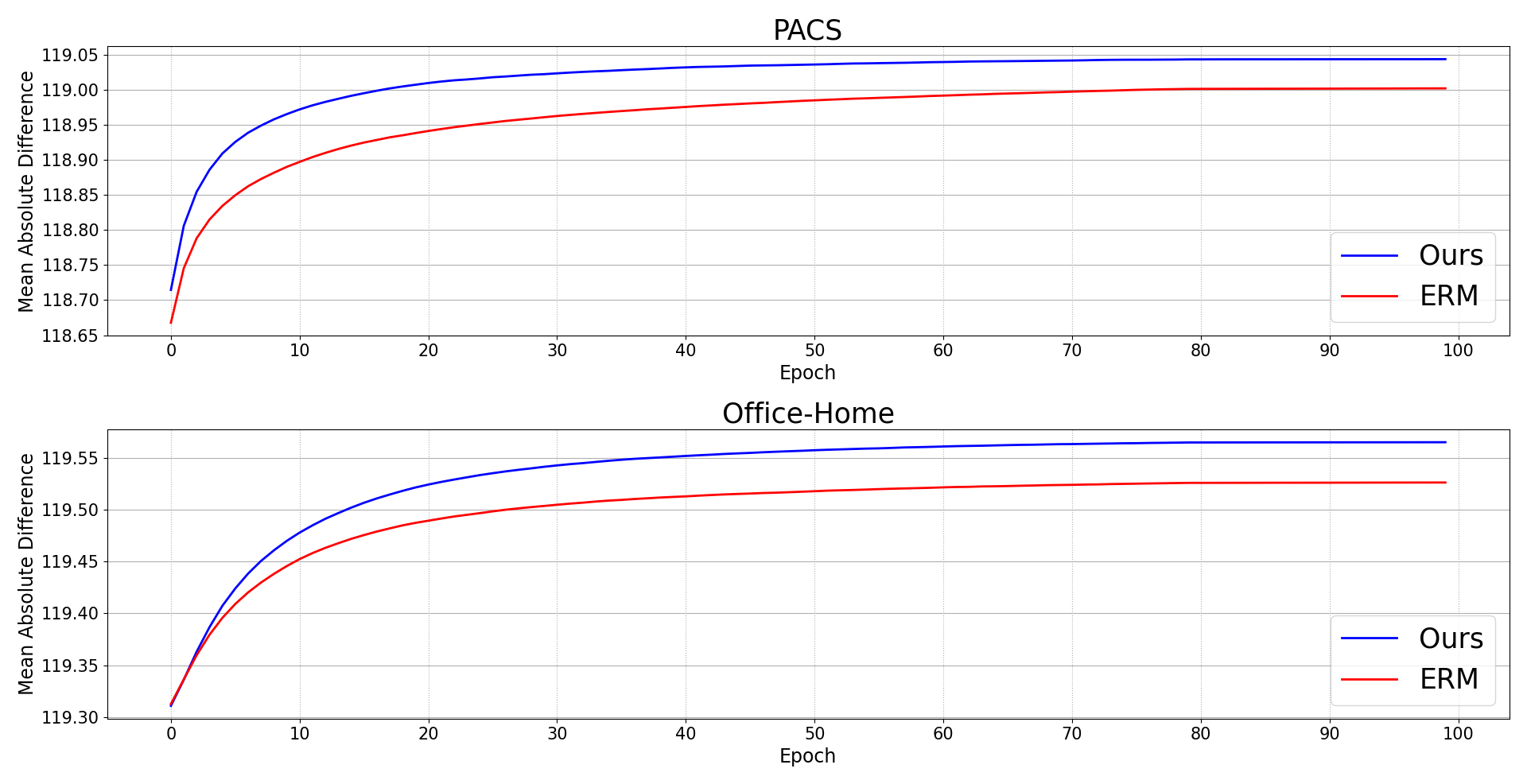}
\caption{Mean absolute difference for ERM and our approach.}
\label{fig:diversity_metric}
\end{figure*}

\begin{table*}
\begin{center}
% \scriptsize
\begin{tabular}{|c|c|c|c|c|}
\hline
  & \multicolumn{2}{|c|}{\textit{without TTBN}} & \multicolumn{2}{|c|}{\textit{with TTBN}}\\
\hline
Ablation & Avg. Val. Acc. & Avg. Test Acc. & Avg. Val. Acc. & Avg. Test Acc.\\
\hline
\multicolumn{5}{|c|}{\textbf{PACS dataset}}\\
\hline
Double branch only (A) & $ 96.8 \pm  0.6$ & $ 53.4 \pm  2.6$ & $ 96.4 \pm  0.3$ & $ 67.4 \pm  0.8$\\
\hline
Detached loss term (B) & $ 97.5 \pm  0.1$ & $ 52.6 \pm  2.3$ & $ 97.3 \pm  0.3$ & $ 68.2 \pm  1.3$ \\
\hline 
Secondary prediction branch (C) & $ 98.0 \pm  0.1$ & $ 53.4 \pm  2.8$ & $ 96.9 \pm  0.2$ & $ 70.1 \pm  0.4$ \\
\hline 
Single branch (D) & $ 92.8 \pm  1.1$ & $ 46.4 \pm  4.9$ & $ 93.0 \pm  0.9$ & $ 51.2 \pm  5.1$ \\
\hline
\multicolumn{5}{|c|}{\textbf{Office-Home dataset}}\\
\hline
Double branch only (A) & $ 82.7 \pm  0.4$ & $ 52.8 \pm  0.5$ & $ 82.6 \pm  0.3$ & $ 53.5 \pm  0.4$ \\
\hline
Detached loss term (B) & $ 83.5 \pm  0.7$ & $ 52.7 \pm  0.6$ & $ 82.3 \pm  0.6$ & $ 54.0 \pm  0.6$ \\
\hline 
Secondary prediction branch (C) & $ 81.3 \pm  0.4$ & $ 53.9 \pm  0.7$ & $ 83.8 \pm  0.6$ & $ 54.8 \pm  0.5$ \\
\hline 
Single branch (D) & $ 82.6 \pm  0.7$ & $ 53.7 \pm  0.4$ & $ 82.0 \pm  0.5$ & $ 54.3 \pm  0.5$ \\
\hline
\end{tabular}
\end{center}
\caption{Ablation Study}
\label{table:ablation_study}
\end{table*}

\begin{table*}
\begin{center}
% \scriptsize
\begin{tabular}{|c|c|c|c|c|c|c|}
\hline
\multicolumn{7}{|c|}{\textbf{Avg. test Acc. on PACS - Avg. test Acc. on Office-Home}}\\
\hline
$lr_{+}$ $\downarrow$ / $\alpha$ $\rightarrow$ & $10^{-3}$ & $10^{-2}$ & $0.1$ & $1.0$ & $10.0$ & $100.0$ \\
\hline
$10^{-3}$ & 66.9 - 53.7 & 66.8 - 53.1 & 67.7 - 53.2 & 67.0 - 53.5 & 67.4 - 53.2 & 68.5 - 53.7 \\
$10^{-2}$ & 67.8 - 53.2 & 67.8 - 53.1 & 67.6 - 53.3 & 67.7 - 52.4 & 68.6 - 53.9 & 70.6 - 53.2 \\
$0.1$ & 67.8 - 53.0 & 67.5 - 53.2 & 67.4 - 53.3 & 69.5 - 53.8 & \textbf{71.3 - 54.7} & 69.2 - 51.9\\
$1.0$ & 67.1 - 53.3 & 68.0 - 53.4 & 69.0 - 53.8 & \textbf{71.3 - 54.4} & 70.3 - 52.6 & 20.1 - 49.9\\
$10.0$ & 67.8 - 52.9 & 67.2 - 53.4 & 67.4 - 53.3 & 66.0 - 53.9 & 54.4 - 51.9 & 15.0 - 5.2\\
$100.0$ & 66.2 - 53.2 & 67.9 - 53.4 & 67.4 - 53.4 & 67.8 - 53.3 & 60.5 - 53.2 & 14.5 - 2.0\\
\hline
\end{tabular}
\end{center}
\caption{Broad hyper-parameters sensitivity analysis.}
\label{table:hparams_sensitivity_analysis}
\end{table*}

We also conducted an extensive ablation study to understand and demonstrate the necessity of our choices. As a sanity check, we first study the $\alpha = 0$ situation: a single features extractor on which two classification layers are plugged in, trained only with the cross-entropy on the same batch at each iteration for both branches (line A in the table \ref{table:ablation_study}). The differences of initialization of the classifiers may have an implicit ensembling effect, as in MIMO \cite{havasi2021training}, which could lead to a better out-of-distribution generalization without the need for the shortcut avoidance loss. We also study the effect of detaching from the computational graph the $c_2(f(W, \tilde{x}_i))$ term (not optimizing the features extractor with respect to this part of the loss) in the shortcut avoidance loss (line B), as this could lead to a substantial improvement in memory consumption, and as the simultaneous optimization on both terms in not needed \textit{per se} to decrease the generalization ability of the network. Then, to show that the performance gain is effectively linked to a mitigation of the shortcut learning phenomenon, we study the impact of using the secondary prediction branch at test-time rather than the primary one (line C) and the effect of applying our shortcut avoidance loss on an architecture without the added secondary branch (line D). To further show the effect of our loss, we track during training a measure of the diversity of the learned patterns for both our approach and ERM. Inspired by \cite{8603826}, we use the mean absolute difference ($MAD$) between normalized convolutional filters $f$ (or neurons for fully connected layers) of a certain layer, computed over all layers $L$ of size $N_L$ and training domains $D$, for an epoch $t$, following the equation \ref{equation:diversity}. The results are available in figure \ref{fig:diversity_metric} and show a systematic increase in the diversity of the learned patterns for our approach compared to ERM, for both benchmarks. Finally, as the tuning of hyper-parameters in the domain generalization setting is a critical issue because the target domain cannot be used, we conduct a broad hyper-parameters sensitivity analysis, available in table \ref{table:hparams_sensitivity_analysis}, showing a relatively low sensitivity and a large match between hyper-parameters fit for all training-test pairs of PACS and Office-Home.

\begin{equation}
    MAD(t) = \sum_{D} \sum_{L} \frac{1}{{N_L}^2}\sum_{i,j} ||f_{t, D, L, i} - f_{t, D, L, j}||_1
\label{equation:diversity}
\end{equation}

The results of the ablation study outline several things: using two prediction branches without the additional loss yield a small increase of performance on both benchmarks, but it remains far below our approach whose gain is therefore not coming from an implicit ensembling mechanism. Detaching the first half of the shortcut avoidance loss from the computational graph shows a decreased performance as well. This detachment most likely only results in a slower learning as the constraint's gradient pushes in the reverse direction of the classification loss gradient. This behavior is prevented when the features extractor is optimized with regard to both terms of the regularization: pushing in the reverse direction of the classification gradient will only slide the difference in the parameter space, but not shorten the gap. The use of the secondary prediction branch at test-time results in performance fairly similar than the first branch, only lower in validation. This was to be expected as the secondary branch is precisely trained so that it generalizes less on the training domain. Finally, the use of our shortcut avoidance loss applied on the original model (no added prediction branch) results in a dramatic drop in accuracy on the PACS dataset but not on the Office-Home dataset, most likely due to the higher diversity in Office-Home that prevents the original patterns from being ignored.

\section{Conclusion}

In this paper, we investigated the behavior of different single-source methods when used in conjunction with test-time batch normalization, on the PACS and Office-Home benchmarks. We showed that test-time batch normalization always have a positive, yet highly variable, influence, and that most of the time the addition of a training-time method is superfluous. We hypothesized that this lack of additional performance was linked to the selection behavior of some algorithms, which still learn the same subset of patterns as the standard training, but weight them differently. We finally proposed a novel approach learning normally "hidden" patterns by looking for predictive patterns that generalize less and showed that it yielded state-of-the-art results on both benchmarks. Future work will be dedicated to a better understanding of the origin of this test-time batch normalization variability and to experiments with our method on the DomainBed \cite{gulrajani2021in} benchmark.

\section*{Acknowledgement}

This work was in part supported by the 4D Vision project funded by the Partner University Fund (PUF), a FACE program, as well as the French Research Agency, l’Agence Nationale de Recherche (ANR), through the projects Learn Real (ANR-18-CHR3-0002-01), Chiron (ANR-20-IADJ-0001-01), Aristotle (ANR-21-FAI1-0009-01), and the joint support of the French national program of investment of the future and the regions through the PSPC FAIR Waste project.

%-------------------------------------------------------------------------
\newpage
{\small
\bibliographystyle{ieee_fullname}
\bibliography{egbib}
}

\end{document}